\documentclass[11pt,a4paper]{article}
\usepackage[hyperref]{acl2018}
\usepackage{times}
\usepackage{latexsym}
\usepackage{lipsum}
\usepackage{graphicx}
\usepackage{float}
\usepackage{url}
\usepackage{textcomp}
\usepackage{booktabs}
\usepackage{tabularx}
\usepackage{array}
\usepackage{tikz}
\usepackage{caption}
\usetikzlibrary[arrows.meta,bending]
\usetikzlibrary{positioning}
\usepackage{graphicx}
\graphicspath{{images/}}

\newcolumntype{L}[1]{>{\raggedright\let\newline\\\arraybackslash\hspace{0pt}}m{#1}}
\newcolumntype{C}[1]{>{\centering\let\newline\\\arraybackslash\hspace{0pt}}m{#1}}
\newcolumntype{R}[1]{>{\raggedleft\let\newline\\\arraybackslash\hspace{0pt}}m{#1}}

\aclfinalcopy 

\newtheorem{thm}{Theorem}[section] 

\newtheorem{defn}[thm]{Definition} 

\title{Privacy-preserving Machine Learning for Medical Image Classification}

\author{Shreyansh Singh and K.K. Shukla\\ \\
  Indian Institute of Technology (BHU), Varanasi, India \\
  {\tt \{shreyansh.singh.cse16,kkshukla.cse\}@iitbhu.ac.in} \\}

\date{}

\begin{document}
\maketitle
\begin{abstract}
  With the rising use of Machine Learning (ML) and Deep Learning (DL) in various industries, the medical industry is also not far behind. A very simple yet extremely important use case of ML in this industry is for image classification. This is important for doctors to help them detect certain diseases timely, thereby acting as an aid to reduce chances of human judgement error. However, when using automated systems like these, there is a privacy concern as well. Attackers should not be able to get access to the medical records and images of the patients. It is also required that the model be secure, and that the data that is sent to the model and the predictions that are received both should not be revealed to the model in clear text.
  
  In this study, we aim to solve these problems in the context of a medical image classification problem of detection of pneumonia by examining chest x-ray images.
  
\end{abstract}

\section{Introduction}

The use of automation in the form of Machine Learning is becoming more visible in the medical industry by the day. However, letting the computers handle confidential medical information of the patients comes with the danger of attackers getting access to those and probably misusing them. Hence, security and privacy should be an important point of concern when setting up such a system. \\
To address this, privacy-preserving machine learning (or private machine learning) was introduced. Private machine learning is a combination of cryptography, machine learning and distributed systems. This requires a deep understanding of cryptography concepts like Secure Multiparty Computation, Differential Privacy and most importantly how these can be integrated with machine learning to perform privacy-preserving predictions.

\subsection{Secure Multiparty Computation}
Secure multi-party computation (SMPC) also known as secure computation is a sub-field of cryptography where the goal is to create a provision for parties to jointly compute a function over their inputs, which are kept private i.e. not shared with the other parties. This model is different from traditional cryptography because of the fact that here the information is to be protected from the other participants instead of from an adversary who is outside the system. A basic understanding of SMPC can be obtained from Shamir’s Secret Sharing Scheme ~\cite{Shamir}. The purpose of that scheme is to divide and distribute one secret value among the participants. A subset of the participants must pool their data to retrieve the secet value.
Shamir’s scheme can also be used on a secret shared value to perform some computation. The result of every participant’s computation (on their own data) can be grouped together to get the required outcome without ever revealing the secret inputs.

The definition of an MPC task involves defining the following -

\begin{itemize}
    \item \textbf{Functionality} - What is needed to be computed?
    \item \textbf{Security type} - How strong of a protection is required?
    \item \textbf{Adversarial model} - What do we want to protect against?
    \item \textbf{Network model} - In which setting will it be done?
\end{itemize}

The functionality is the code of the trusted party. Security type is of 3 types - \textit{Computational, Statistical} and \textit{Perfect}. The Adversarial model can be described in different ways - 
\begin{itemize}
    \item \textbf{Adversarial behavior}
    \begin{itemize}
        \item \textit{Semi honest} - corrupted parties follow the protocol honestly, but an external adversary \textit{A} tries to learn more information. Also called honest-but-curious. 
        \item \textit{Fail stop} - same as semi honest, but corrupted parties can prematurely halt.
        \item \textit{Malicious} - corrupted parties can deviate from the protocol in an arbitrary way
    \end{itemize}
    \item \textbf{Adversarial power}
    \begin{itemize}
        \item \textit{Polynomial time} - computational security, normally requires cryptographic assumptions. E.g., encryption, signatures, oblivious transfer
        \item \textit{Computationally unbounded} - an all-powerful adversary, information-theoretic security
    \end{itemize}
    \item \textbf{Adversarial corruption}
    \begin{itemize}
        \item \textit{Static} - the set of corrupted parties is defined before the execution of the protocol begins. Honest parties are always honest, corrupted parties are always corrupted
        \item \textit{Adaptive} - Adversary \textit{A} can decide which parties to corrupt during the course of the protocol, based on information it dynamically learns
        \item \textit{Mobile} - Adversary \textit{A} can ``jump'' between parties. Honest parties  can become corrupted, corrupted parties can become honest again
    \end{itemize}
    \item \textbf{Number of corrupted parties}
    \begin{itemize}
        \item Number of corrupted parties, \textit{t} is governed by the condition, $t \leq n$ denoting an upper bound on the number of corruptions. Types -
        \begin{itemize}
            \item No honest majority, e.g., two-party computation
            \item Honest majority, i.e., $t < n/2$
            \item Two-thirds majority, i.e., $t < n/3$
        \end{itemize}
        \item \textit{General adversary structure} - Protection against specific subsets of parties
    \end{itemize}
\end{itemize}

The communication/network model can of the following types -
\begin{itemize}
    \item \textbf{Point-to-point}: fully connected network of pairwise channels. 
    \begin{itemize}
        \item \textit{Unauthenticated channels}
        \item \textit{Authenticated channels}: in the computational setting
        \item \textit{Private channels}: in the IT setting
    \end{itemize}
    \item \textbf{Broadcast} - additional broadcast channel
\end{itemize}
SMPC gives a combination of encryption, distribution and distributed combination and this has a big impact on data security and data privacy.

\subsection{Differential Privacy}
\label{dp_def}

Differential Privacy (DP) is a rigorous mathematical framework which allows sharing information about a dataset publicly by describing the patterns of groups of the dataset without revealing information about the individuals in the dataset. An algorithm is said to be differentially private if and only if the inclusion of any one instance in the training dataset causes only statistically minor changes to the output of the algorithm. This is required in situations where, for example, the identity of a patient (in the medical context) is to be kept private. If not for DP, and using just the trained ML model, attackers would have been capable of finding out the hospital that a specific patient belonged to, which would violate their right to privacy. The role of DP here is to limit the attacker's ability to infer such membership by putting a theoretical limit on the influence that a single individual can have.

However using DP means that there will be a tradeoff between accuracy and security. Although the aim of DP is to minimise the "information leak" from a single query, but keeping this value small enough when multiple queries are made can become a challenge as for every query, the total ``information leak'' will increase. As a solution, more noise has to be injected in the data to minimise the privacy leakage but that would mean the accuracy of the model will go down. This can be a big problem when training complex ML models.

\section{Literature Review}

The literature review was conducted using exhaustive search over the following terms: ``secure multiparty computation'', ``differential privacy'', ``privacy preserving machine learning'' and ``secure deep learning''. Apart from keyword search and relevance, other selection criteria were the chronology of the papers and the quality of sources (peer reviewed journals and conferences).

\subsection{Secure Computation}

Secure Multiparty computation (SMC) can be divided into two broad classes - Two-party computation and Multi-party computation.

\subsubsection{Two-party computation}
~\cite{Yao} first introduced the idea of two-party computation (2PC). The idea of Yao's garbled circuits were introduced in ~\cite{Yao_use} although it was heavily based on ~\cite{Yao}. Yao's garbled circuits facilitates two-party secure computation in which two mistrusting parties can jointly evaluate a function over their private inputs without the presence of a trusted third party. Yao's basic protocol is secure against semi-honest adversaries. 2PC protocols in a malicious setting (secure against active adversaries) were proposed a bit later in ~\cite{2PC_Malicious}, ~\cite{OT_Efficiently} and ~\cite{Two_party_LEGO}. A solution which works with committed inputs explicitly was given by ~\cite{CommitedInp_Malicious}.

\subsubsection{Multi-party computation}
Secret sharing forms the fundamentals of multi-party computation (MPC). The two most commonly used methods are Shamir's secret sharing ~\cite{Shamir} and additive secret sharing. There has been a lot of work on using MPC with secret sharing schemes. One of the most popular is SPDZ ~\cite{SPDZ}. This uses additive secret shares and is secure against active adversaries (malicious, dishonest majority). Some other implementations of secure MPC protocols exist like ~\cite{ABY}, ~\cite{OblivC}, ~\cite{SCALE-MAMBA} and  ~\cite{FRESCO}. These however are independent frameworks that do not help much in Machine Learning in terms of integration with current ML platforms and that they simply provide implementations of the SMC protocols rather than focus on private machine learning. ~\cite{securenn} is an SMC framework which provides efficient 3-party protocols tailored for state-of-the-art neural networks. Other such frameworks include SecureML ~\cite{secureml}, GAZELLE ~\cite{gazelle} and ABY3 ~\cite{aby3}. These frameworks focus on adapting secure computation protocols to private machine learning. Crypten (~\cite{crypten}) is a recent framework developed by Facebook Research for privacy preserving machine learning on Pytorch but is still quite limited in terms of the features it offers from the deep learning perspective. TF encrypted ~\cite{tfencrypted} is a framework which provides secure multi-party computation directly in TensorFlow. We use TF encrypted as the framework for SMC in our research.

\subsection{Differential Privacy}

Differential Privacy (DP), as described in \ref{dp_def}, is a rigorous mathematical framework which allows sharing information about a dataset publicly by describing the patterns of groups of the dataset without revealing information about the individuals in the dataset. In ~\cite{hospital-attack} the authors showed that if, for example, an attacker gets access to an ML model being used in a hospital dealing with private medical information of patients, the attackers can infer whether an individual was a patient at the hospital or not, thus violating their right to privacy. DP can be formally stated as in ~\cite{dp-def} - 

\begin{defn}
    A randomized mechanism K provides ${(\epsilon, \delta)}$ - differential privacy if for any two neighboring database $D_{1}$ and $D_{2}$ that differ in only a single entry, ${\forall S \subseteq Range(K)}$,
    \begin{equation}
       Pr(K(D_1) \in S)  \leq  e^{\epsilon}Pr(K(D_2) \in S) + \delta
    \end{equation}
\end{defn}

If ${\delta = 0}$, ${K}$ is said to satisfy ${\epsilon}$-differential privacy.

The idea is that to achieve DP, noise is added to the algorithm's output. This noise is dependent on the sensitivity of the output, where sensitivity is the measure of the maximum change of output due to the inclusion of a single data instance ~\cite{federated_learning}.

Two popular mechanisms for achieving DP are the Laplacian and Gaussian mechanisms. When an algorithm requires multiple additive noise mechanisms,the evaluation of the privacy guarantee follows from the basic composition theorem ~\cite{dp_1}, ~\cite{dp_2} or from advanced composition theorems and their extensions ~\cite{conc_privacy}, ~\cite{conc_privacy2}.

As a tool that could integerate the use of Differential Privacy with Machine Learning, Tensorflow Privacy (~\cite{tf_privacy}) was introduced. Tensorflow Privacy is a library that includes implementations of TensorFlow optimizers for training machine learning models with differential privacy.

\section{Data}

Medical image classification includes a vast array of problems to work on. This research takes up the task of detecting pneumonia in patients by analysing their chest X-ray images. The dataset is obtained from ~\cite{dataset_orig}, a research published in the scientific journal \textit{Cell}, where the authors have collected such medical images and aim to apply image-based deep learning to detect such diseases. A copy of the dataset is also available on Kaggle ~\cite{dataset_kaggle}.

\begin{figure}[!h]
    \begin{center}
        \includegraphics[width=\linewidth]{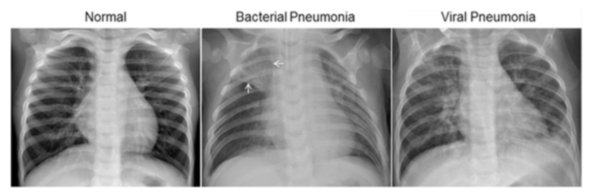}
        \caption{Illustrative Examples of Chest X-Rays in Patients with Pneumonia, ~\cite{dataset_orig}}
        \label{fig:dataset_example}
  \end{center}
\end{figure}

Figure \ref{fig:dataset_example} shows the variation in the x-rays for the different kinds of pneumonia. According to ~\cite{dataset_orig}, the normal chest X-ray (left panel) depicts clear lungs without any areas of abnormal opacification in the image. Bacterial pneumonia (middle) typically exhibits a focal lobar consolidation, in this case in the right upper lobe (white arrows), whereas viral pneumonia (right) manifests with a more diffuse ``interstitial'' pattern in both lungs.

\subsection{Description of the dataset}

The dataset is organised into 3 folders (train, test, val) and contains sub folders for each image category (Pneumonia/Normal). There are 5847 X-ray images (all in JPEG format) and 2 categories (Pneumonia/Normal). 

Chest X-ray images (anterior-posterior) were selected from retrospective cohorts of pediatric patients of one to five years old from Guangzhou Women and Children's Medical Center, Guangzhou. All chest X-ray imaging was performed as part of patients' routine clinical care.

For the analysis of chest X-ray images, all chest radiographs were initially screened for quality control by removing all low quality or unreadable scans. The diagnoses for the images were then graded by two expert physicians before being cleared for training any AI system. In order to account for any grading errors, the evaluation set was also checked by a third expert. ~\cite{dataset_orig}

The type of the image, i.e. Pneumonia or Normal can be identified from the filename as well. The Normal images have the ``NORMAL'' keyword in them. The ones indicating Pneumonia have the keywords ``virus'' or ``bacteria'' in them indicating the type. In our experiments however, we haven't considered the type of Pneumonia in the classification. It is just Normal X-ray images or the ones with Pneumonia. 

Breakdown of the data for classification - 
\begin{itemize}
    \item \textbf{Train data} - 1341 normal images and 3882 pneumonia images
    \item \textbf{Test data} - We merge the images in the test and val folders because there were only 18 images in the test folder. After merging, there were 234 normal images and 390 pneumonia images.
\end{itemize}

\subsection{Data Preprocessing}
We have done two different types of preprocessing of the images in order to perform different experiments.

\subsubsection{Preprocessing Technique 1}
For each image in the training and testing data, we perform the following steps - 
\begin{enumerate}
    \item Resize the image into size (125, 150) using \textit{resize} function of the Pillow library
    \item Convert the image into greyscale
    \item Divide each pixel value by 255
    \item Save each image as a numpy array along with their labels (0 or 1 for Normal or Pneumonia respectively)
\end{enumerate}

\subsubsection{Preprocessing Technique 2}
For each image in the training and testing data, we perform the following steps - 
\begin{enumerate}
    \item Resize the image into size (224, 224) using \textit{cv2.INTER\_CUBIC}, i.e. cubic interpolation using the OpenCV library in Python
    \item Convert the image into a numpy array and also store the corresponding labels
\end{enumerate}

We save both these arrays as pickle files to be later used with our deep learning models.

\section{Approach}

Our approach can be described in 3 steps, model training, model serving and serving the model predictions. They are described below - 

\subsection{Model Training}

The results for both the approaches are given in Section~\ref{results_sec}.

The first step of our approach is the model training. We have to train a Deep Learning model to do the image classification task for us. We used Tensorflow as the Deep Learning framework and used Keras for training our deep learning models. The model training was done on Google Colab using the Tesla P100 GPU and 25 GB of RAM. We trained our model both with and without differential privacy. 

In all, we used four models for our experiments. The first two are custom made image classification models. The first of the two used \textit{AveragePooling2D} while the other uses \textit{MaxPooling2D} as the pooling layer. Both of these accept images preprocessed using the first preprocessing method. The third and fourth models are based on the VGG16 architecture, on of which accepts images preprocessed using the first preprocessing method while the other accepts the images preprocessed using the second preprocessing technique.
In order in which the models were described, we will name them DNN-Av, DNN-Max, VGG16-1, VGG16-2.

For training with differential privacy, we used Tensorflow Privacy ~\cite{tf_privacy}, specifically the \textit{DPGradientDescentGaussianOptimizer}. This is an extension of the Stochastic Gradient Descent ~\cite{sgd_orig} Optimizer. Certain modifications to the SGD optimizer, like the following, can make it differentially  private.
\begin{itemize}
    \item The sensitivity of each gradient needs to be bounded. The amount of influence each individual training point which is sampled in a minibatch, can create on the the resulting gradient computation, needs to be limited. This can be done by clipping the gradient computed on each training point with respect to the model parameters. This clipped gradient should be used with the learning rate to update the model parameters. This keeps a check on the amount that each training  point can impact the model parameters.
    \item Some randomness needs to be introduced in the algorithm's behavior to make it very difficult statistically to to know whether or not a particular point was included in the training set by comparing the updates stochastic gradient descent applies when it operates with or without this particular point in the training set. This can be achieved by sampling random noise and adding it to the clipped gradients.
\end{itemize}

\textit{DPGradientDescentGaussianOptimizer} does exactly this. Apart from the learning rate, population size (basically the size of the training set). The population size helps in calculating the strength of privacy achieved. Three new hyperparmeters are also passed to this optimizer, namely \textit{l2\_norm\_clip}, \textit{noise\_multiplier}, \textit{num\_microbatches}. \textit{l2\_norm\_clip} is the maximum Euclidean norm of each individual gradient computed on an individual example from a minibatch. \textit{noise\_multiplier} controls how much noise is sampled and added to gradients before they are applied by the optimizer. More noise usually provides a better security but at the same time reduces the accuracy, which is a tradeoff. Microbatches, explained in ~\cite{tf_privacy}, introduce a new granularity by splitting each minibatch into multiple microbatches. With this, the clipping of gradients need not be done on a per-example basis but can now be done in microbatches. So, here we have a tradeoff between performance (small value of the parameter) and utility (large value of the parameter).

The use of the different preprocessing methods and model combinations are just to understand which method and which model works best for us.

\subsection{Model Serving}

After training the model with Keras, we now have to secure the model and serve it. For this we have used the tf-encrypted ~\cite{tfencrypted} framework. In this stage, tf-encrypted (TFE) helps to set-up and perform MPC on the model. We use Secure-NN ~\cite{securenn} as the algorithm to perform MPC. For this, we set-up three local TFE servers. The idea here is that the model weights and input to the model are split and a share of each value is sent to the servers. The important thing to note here is that if one looks at the share on one server, no information is revealed about the original values i.e. the input data or the model weights. TFE creates a clone of the model suited for performing MPC. 

After this, a queuing server is set up that allows the TFE servers to wait for and accept prediction requests for the secured model by external clients. Upon receiving such prediction requests, the \textit{QueueServer} is responsible to serve predictions to the client. The configuration of the queue server is stored in a file which will be read by the client (explained in Section ~\ref{servepred_section}).

\subsection{Serving the private predictions}
\label{servepred_section}
In parallel with the model serving service, we also set up a \textit{QueueClient}, this queue is to request the private predictions from the model. This client connects to the queue server using the same configuration as the server. The queue is responsible for sharing the plaintext data secretly before submitting the shares in a prediction request. For querying the model, the image is inserted into the queue, the data is shared between the TFE servers locally. These shares are then sent to the queuing serve to obtain predictions, as explained in the previous section.

Finally, what we have, is a pipeline which allows us to train our model, serve it, and make predictions through it. The key point is that the predictions do not reveal any private information to the service at any point. The model host never sees the input data or the predictions and the model is never downloaded. We are able to get the private predictions on encrypted data with an encrypted model.

\section{Results}
\label{results_sec}
\section{Model Performance}

The accuracy and loss of the models on the test dataset is shown below -
\begin{table}[!h]
    \centering
    \begin{tabular}{| c | c | c | c |} 
     \hline
     Model & Accuracy & Loss\\ [0.5ex] 
     \hline\hline
     DNN-Av &  0.852 & 0.33 \\
     \hline
     DNN-Max &  0.873 & 0.29 \\
     \hline
     VGG16-1 & 0.809 & 0.53 \\
     \hline
     VGG16-2 & 0.964 & 0.993 \\
    \hline
    \end{tabular}
    \caption{Test set results for 2-class classification}
    \label{tab:classifier_results}
\end{table}

When trained with Differential Privacy, the model accuracies are shown below - 

\begin{table}[!h]
    \centering
    \begin{tabular}{| c | c |} 
     \hline
     Model & Accuracy \\ [0.5ex] 
     \hline\hline
     DNN-Av & 0.721 \\
     \hline
     DNN-Max &  0.748 \\
     \hline
     VGG16-1 &  0.705 \\
     \hline
     VGG16-2 &  0.853 \\
    \hline
    \end{tabular}
    \caption{Test set results for 2-class classification when training with Differential Privacy}
    \label{tab:classifier_results}
\end{table}

The plot of accuracy-vs-epoch and loss-vs-epoch of each of the models are also shown below - 
\begin{figure}[!h]
\centering
\begin{minipage}{.25\textwidth}
  \centering
  \includegraphics[width=\linewidth]{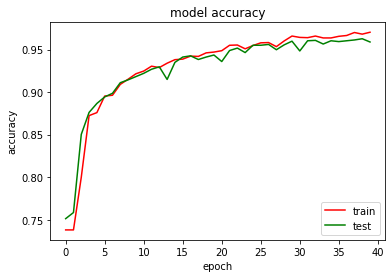}
  \caption{DNN-Av - Accuracies}
  \label{fig:avepool_acc}
\end{minipage}%
\begin{minipage}{.25\textwidth}
  \centering
  \includegraphics[width=\linewidth]{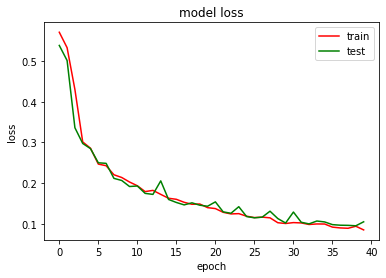}
  \caption{DNN-Av - Losses}
  \label{fig:avepool_loss}
\end{minipage}
\end{figure}

\begin{figure}[!h]
\centering
\begin{minipage}{.25\textwidth}
  \centering
  \includegraphics[width=\linewidth]{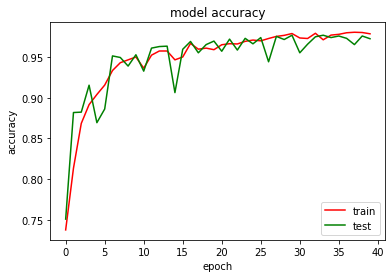}
  \caption{DNN-Max - Accuracy}
  \label{fig:maxpool_acc}
\end{minipage}%
\begin{minipage}{.25\textwidth}
  \centering
  \includegraphics[width=\linewidth]{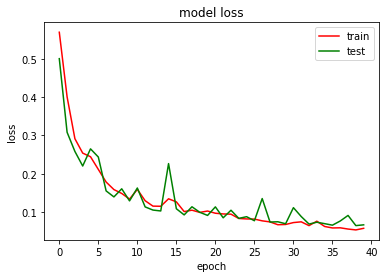}
  \caption{DNN-Max - Loss}
  \label{fig:maxpool_loss}
\end{minipage}
\end{figure}

\begin{figure}[!h]
\centering
\begin{minipage}{.25\textwidth}
  \centering
  \includegraphics[width=\linewidth]{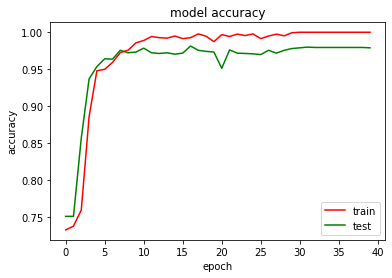}
  \caption{VGG16-1 - Accuracy}
  \label{fig:vgg161_acc}
\end{minipage}%
\begin{minipage}{.25\textwidth}
  \centering
  \includegraphics[width=\linewidth]{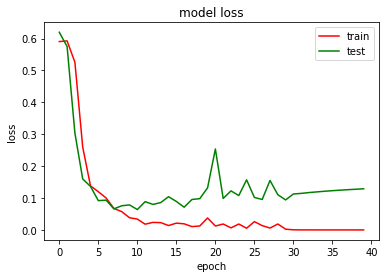}
  \caption{VGG16-1 - Loss}
  \label{fig:vgg161_loss}
\end{minipage}
\end{figure}

\begin{figure}[!h]
\centering
\begin{minipage}{.25\textwidth}
  \centering
  \includegraphics[width=\linewidth]{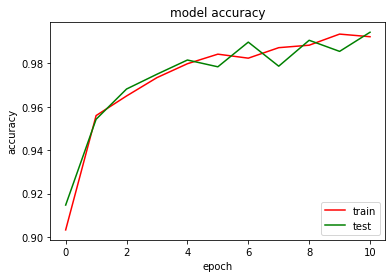}
  \caption{VGG16-2 - Accuracy}
  \label{fig:vgg162_acc}
\end{minipage}%
\begin{minipage}{.25\textwidth}
  \centering
  \includegraphics[width=\linewidth]{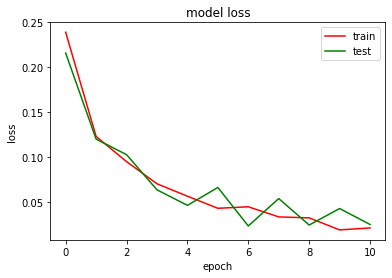}
  \caption{VGG16-2 - Loss}
  \label{fig:vgg162_loss}
\end{minipage}
\end{figure}

For the models trained with DP, the accuracies and losses vary a lot with epochs. We trained DNN-Av, DNN-Max and VGG16-1 with DP, and the plots are shown below - 

\begin{figure}[!h]
\centering
\begin{minipage}{.25\textwidth}
  \centering
  \includegraphics[width=\linewidth]{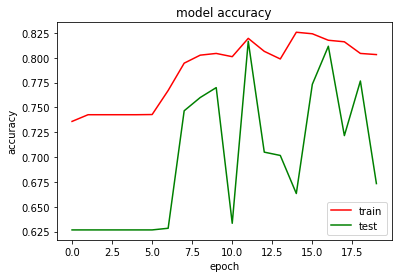}
  \caption{DNN-Av (with DP) - Accuracies}
  \label{fig:avepool_dp_acc}
\end{minipage}%
\begin{minipage}{.25\textwidth}
  \centering
  \includegraphics[width=\linewidth]{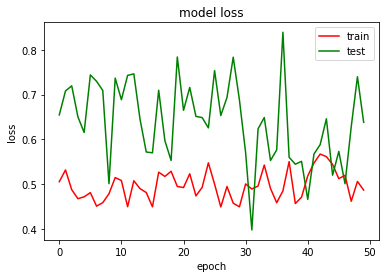}
  \caption{DNN-Av (with DP) - Losses}
  \label{fig:avepool_dp_loss}
\end{minipage}
\end{figure}

\begin{figure}[!h]
\centering
\begin{minipage}{.25\textwidth}
  \centering
  \includegraphics[width=\linewidth]{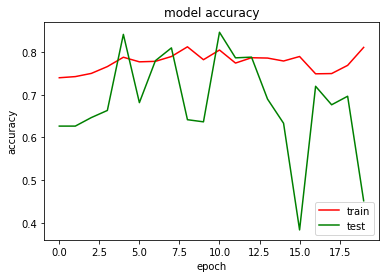}
  \caption{DNN-Max (with DP) - Accuracy}
  \label{fig:maxpool_dp_acc}
\end{minipage}%
\begin{minipage}{.25\textwidth}
  \centering
  \includegraphics[width=\linewidth]{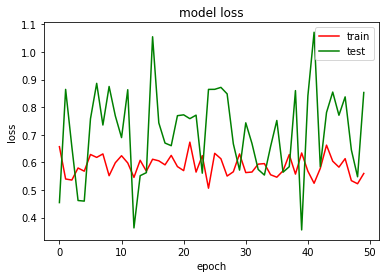}
  \caption{DNN-Max (with DP) - Loss}
  \label{fig:maxpool_dp_loss}
\end{minipage}
\end{figure}

\begin{figure}[!h]
\centering
\begin{minipage}{.25\textwidth}
  \centering
  \includegraphics[width=\linewidth]{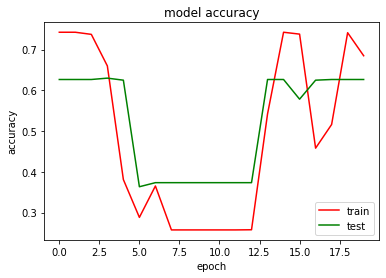}
  \caption{VGG16-1 (with DP) - Accuracy}
  \label{fig:vgg161_dp_acc}
\end{minipage}%
\begin{minipage}{.25\textwidth}
  \centering
  \includegraphics[width=\linewidth]{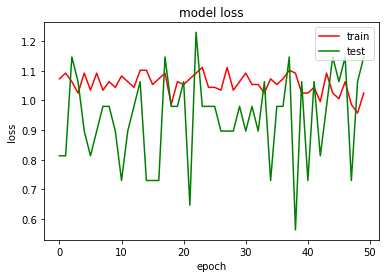}
  \caption{VGG16-1 (with DP) - Loss}
  \label{fig:vgg161_dp_loss}
\end{minipage}
\end{figure}

\section{Analysis}

\subsection{Model Evaluation on the data}
The VGG16-based model when trained with the images generated using second preprocessing method is the best model in terms of accuracy (96.41\%) among all the other models. However, the VGG16 model trained on the data generated using first preprocessing method performs worse than the other two classes of models. It achieves an accuracy of 80.9\% and a cross entropy loss of 0.53. In case of the other two models, the one with MaxPooling (accuracy = 87.33\%, loss = 0.29) performs better than the model with AveragePooling (accuracy = 85.2\%, loss = 0.33). All these models were trained for 40 epochs with a ModelCheckpoint (in Keras) to save the best model in terms of validation accuracy and EarlyStopping (in Keras) to stop the training when the model performance tends to decrease. Using Differential Privacy (DP) while training the models, results in a reduced accuracy (84.89\%) which should also happen in theory as noise is introduced in the data while training. The accuracy (on an average) reduced by around 12\% when training with DP. The DNN-Av and DNN-Max models were the most affected with reduction in accuracy of 13\% where the VGG16 models saw an accuracy drop of 10\% and 11\% respectively. 

\subsection{Limitations}
TF Encrypted is a fairly recent framework developed in 2018. When cloning Keras models, TF Encrypted usually works really well, however it has some limitations. Some advanced Keras layers are still not implemented in that library. There were some issues with Batch Normalization and Dropout layers as well, as they are not correctly implemented in TFE and gives error while conversion from Keras models. 
Also, serving predictions from the encrypted models, is much slower than getting predictions from the unencrypted model. Although, this is expected, but for VGG16 it becomes very slow, which becomes very evident if having to be used as a service.

\section{Conclusion}

Our research shows that it is possible to build a system which can help ensure privacy of the users in a very critical setting where confidentiality of the information is of utmost importance. Private medical data is usually sensitive information which the patient cannot afford to lose. We developed a system which provides private predictions on a Deep Learning model in which the both the model and the data is encrypted and shared (using MPC), for an image classification problem. At no stage of the prediction process, the user is sharing raw data which can be sniffed by the attacker. Furthermore, if we also use Differential Privacy, we ensure that the model does not memorize sensitive information about the training set. This means that it should not be possible for the attackers to reveal some private information by just querying the deployed model. If the model is not trained using DP, the model is vulnerable to attacks such as ~\cite{membership_attack} and ~\cite{model_inversion}, which can help attackers get information about the dataset and in this case, medical information about the patient.

\section{Future Work}
The pipeline described in our work can be used to make a generic framework to provide privacy-preserving medical image classification, i.e. not just limiting to pneumonia detection. 

Also, there were certain limitations of using TF Encrypted for building more complex models as it currently does not support all the layers implemented in Keras. However, TF Encrypted is also open source which means that we can contribute to adding/correcting those layers ourselves. Similar improvements can be done for Crypten ~\cite{crypten} which is also at a very nascent stage in terms of development.

Furthermore, we can also experiment with more models based on different use cases to improve the accuracy and make the prediction time faster (in a setup similar to the one described).

\bibliography{acl2018}
\bibliographystyle{acl_natbib}

\end{document}